\documentclass[sigconf]{acmart}
\copyrightyear{2024}
\acmYear{2024}
\setcopyright{acmlicensed}
\acmConference[ASE '24]{39th IEEE/ACM International Conference on Automated Software Engineering}{October 27-November 1, 2024}{Sacramento, CA, USA}
\acmBooktitle{39th IEEE/ACM International Conference on Automated Software Engineering (ASE '24), October 27-November 1, 2024, Sacramento, CA, USA}
\acmDOI{10.1145/3691620.3695060}
\acmISBN{979-8-4007-1248-7/24/10}

\usepackage{threeparttable}
\usepackage{natbib}
\usepackage{hyperref}
\usepackage{graphicx}
\usepackage{textcomp}
\usepackage{shortcuts}
\usepackage{framed}
\usepackage{xcolor}
\usepackage{url}

\usepackage{breakurl}
\usepackage{multirow}

\usepackage{makecell}
\usepackage{tcolorbox}
\usepackage{enumitem}
\usepackage{rotating}
\usepackage{amsmath}
\usepackage{algorithmicx}

\usepackage[linesnumbered,ruled]{algorithm2e}



\begin{document}


\title{GlitchProber: Advancing Effective Detection and Mitigation of Glitch Tokens in Large Language Models}

\author{Zhibo Zhang}
\authornote{Co-first author with equal contribution.}
\orcid{0009-0008-6447-1756}
\affiliation{%
  \institution{Huazhong University of Science and Technology}
  \city{Wuhan}
  \country{China}
}
\email{zhangzhibom@hust.edu.cn}

\author{Wuxia Bai}
\authornotemark[1]
\orcid{0009-0009-5332-9890}
\affiliation{%
  \institution{Huazhong University of Science and Technology}
  \city{Wuhan}
  \country{China}
}
\email{wuxiabai@hust.edu.cn}

\author{Yuxi Li}
\authornotemark[1]
\orcid{0009-0008-8032-3841}
\affiliation{%
  \institution{Huazhong University of Science and Technology}
  \city{Wuhan}
  \country{China}
}
\email{yuxili@hust.edu.cn}

\author{Mark Huasong Meng}
\orcid{0000-0003-1039-2151}
\affiliation{%
  \institution{Technical University of Munich}
  \city{Munich}
  \country{Germany}
}
\email{huasong.meng@gmail.com}

\author{Kailong Wang}
\authornote{Corresponding Author.}
\orcid{0000-0002-3977-6573}
\affiliation{%
  \institution{Huazhong University of Science and Technology}
  \city{Wuhan}
  \country{China}
}
\email{wangkl@hust.edu.cn}

\author{Ling Shi}
\orcid{0000-0002-2023-0247}
\affiliation{%
  \institution{Nanyang Technological University}
  \city{Singapore}
  \country{Singapore}
}
\email{ling.shi@ntu.edu.sg}

\author{Li Li}
\affiliation{%
  \institution{Beihang University}
  \city{Beijing}
  \country{China}
}
\email{lilicoding@ieee.org}

\author{Jun Wang}
\affiliation{%
  \institution{Beihang University}
  \city{Beijing}
  \country{China}
}
\email{junwang.lu@gmail.com}

\author{Haoyu Wang}
\orcid{0000-0003-1100-8633}
\affiliation{%
  \institution{Huazhong University of Science and Technology}
  \city{Wuhan}
  \country{China}
}
\email{haoyuwang@hust.edu.cn}

\begin{abstract}
Large language models (LLMs) have achieved unprecedented success in the field of natural language processing. However, the black-box nature of their internal mechanisms has brought many concerns about their trustworthiness and interpretability. 
Recent research has discovered a class of abnormal tokens in the model's vocabulary space and named them ``glitch tokens''. 
Those tokens, once included in the input, may induce the model to produce incorrect, irrelevant, or even harmful results, drastically undermining the reliability and practicality of LLMs. 

In this work, we aim to enhance the understanding of glitch tokens and propose techniques for their detection and mitigation. We first reveal the characteristic features induced by glitch tokens on LLMs, which are evidenced by significant deviations in the distributions of attention patterns and dynamic information from intermediate model layers. Based on the insights, we develop \tool, a tool for efficient glitch token detection and mitigation. \tool utilizes small-scale sampling, principal component analysis for accelerated feature extraction, and a simple classifier for efficient vocabulary screening. Taking one step further, \tool rectifies abnormal model intermediate layer values to mitigate the destructive effects of glitch tokens. Evaluated on five mainstream open-source LLMs, \tool demonstrates higher efficiency, precision, and recall compared to existing approaches, with an average F1 score of 0.86 and an average repair rate of 50.06\%. \tool unveils a novel path to address the challenges posed by glitch tokens and inspires future research toward more robust and interpretable LLMs. Our code is available at \href{https://github.com/LLM-Integrity-Guard/GlitchProber}{https://github.com/LLM-Integrity-Guard/GlitchProber}.


\end{abstract}

\maketitle

\section{Introduction}

In the field of Natural Language Processing~(NLP), large language models~(LLMs) like GPT-4~\cite{achiam2023gpt}, Gemini~\cite{team2023gemini, reid2024gemini}, and Claude 3~\cite{anthropic2024claude3} have demonstrated near-human-level text generation capabilities. Their exceptional performance has led to widespread adoption~\cite{wu2023autogen,topsakal2023creating, 10.1145/3650212.3680383}. When using these models, users provide a prompt, which the model's tokenizer breaks down into a series of discrete tokens. These tokens are the fundamental units of information processing for the model, playing a crucial role in the usage of LLMs. Recent research~\cite{SolidGoldMagikarp2023, SolidGoldMagikarpII2023, SolidGoldMagikarpIII2023, geiping2024coercing, petertodd_phenomenon2023, 2023_Glitch_Tokens}, however, has shown that some ``glitch tokens'' exist in the vocabulary of LLMs. Once included in a prompt, these special tokens can potentially lead to model errors, such as misunderstanding user intent, refusing to answer, or generating irrelevant or harmful text. Therefore, thorough analysis and detection of these glitch tokens are crucial to ensure the reliability and safety of LLMs.


To tackle the glitch tokens issue, one notable method recently is presented by Li et al.~\cite{li2024glitch}. This typical and intuitive solution involves studying the characteristics of glitch tokens in the word embedding space of LLMs and accordingly developing detection techniques. They discovered that glitch tokens tend to cluster in the embedding space and proposed an iterative clustering-based technique called \textsc{GlitchHunter} for efficient glitch token detection.

Although there has been progress in detecting glitch tokens, there still lacks an efficient and precise detection of glitch tokens universally applicable in different LLMs. Furthermore, existing approaches primarily focus on detection, however, how to fix the issues caused by glitch tokens in the usage of LLMs remains an open question. Several limitations contribute to the aforementioned challenges:
\begin{enumerate}[left=0pt]
    \item The exhaustive search method of checking vocabulary is simple and intuitive but incurs significant time costs with large token sets, making it inefficient for practical use.
    \item Existing detection methods primarily identify glitch tokens based on features like word frequency and word vectors. However, these features do not deeply explore the mechanisms by which glitch tokens impact model behaviors, resulting in poor detection accuracy and generalization performance.
    \item Current research primarily focuses on detecting glitch tokens rather than how to fix them. While detection can identify issues, it does not eliminate the negative impact of glitch tokens on model performance, limiting its practical value.
\end{enumerate}

\textbf{Our work.} To address these existing challenges and bridge the gap, in this work, we investigate the internal structure of LLMs to explore the differences between glitch tokens and normal tokens. 
Specifically, through empirical study, we discovered significant differences between glitch tokens and normal tokens in terms of the attention patterns and dynamic information of multi-layer perceptron (MLP) modules within transformer-based LLMs. 
This discovery reveals the adversarial impact of glitch tokens on the internal mechanisms of the model, indicating that glitch tokens introduce abnormal interference and noise to neural networks. 
This hinders the model from correctly understanding and processing the semantic information carried by these tokens, ultimately leading to erroneous outputs.

From these findings, we gain an insight that the glitch tokens can be efficiently detected due to the deviated distributions of intermediate layers' outputs caused by them, and accordingly their impact can be effectively mitigated by proactively rectifying those abnormal outputs.
Based on this insight, we propose a new method for glitch token detection and fix called \tool. 
For glitch token detection, \tool first samples a small subset of manually labeled glitch tokens as the sample set, and extracts the outputs of these tokens from the intermediate layers, specifically, the attention scores of the attention patterns and MLP status.
It then applies Principal Component Analysis (PCA)~\cite{abdi2010PCA} dimensionality reduction to the intermediate layers' outputs and obtains a feature representation matrix for the sample set.
This matrix, along with the corresponding class labels, is used to train a Support Vector Machine (SVM)~\cite{cortes1995support} classifier, which can subsequently be employed for glitch token detection. 
To fix the glitch tokens, \tool analyzes the activation value range of normal tokens in the intermediate MLP status and rectifies the activation states of glitch tokens. Specifically, it aims to adjust the activation patterns of glitch tokens to be closer to those of normal tokens, and thereby minimize their impact on the model's output. Our work is published on our website~\cite{Ours}.

\noindent\textbf{Contributions.} We summarize our key contributions as follows:
\begin{itemize}[left=0pt]
    \item \textbf{Empirical Study Exploring the Internal Impact of Glitch Tokens on LLMs.} We conduct a comprehensive and systematic empirical study on how glitch tokens and normal tokens manifest at the structural level across different LLMs. One of our key findings is that glitch tokens can trigger abnormal values in a model's attention patterns and MLP status.
    \item \textbf{Effective Glitch Token Detection.} Our evaluation on five representative open source LLMs demonstrates that \tool can save approximately 40\% of time in glitch token detection compared to the state-of-the-art approaches. Additionally, \tool exhibits a significant improvement in detection accuracy. 
    \item \textbf{Effective Glitch Token Fixing.} 
    In terms of fix, \tool successfully repairs an average of 7,758 tokens across the five LLMs. It achieves an average repair rate of 50.06\%, significantly outperforming the baseline approach. Our results demonstrate the effectiveness of the proposed fix strategy by adjusting the intermediate values of glitch tokens in the intermediate layers.
\end{itemize}

\textbf{Ethical Consideration. }
In this work, we recognize that glitch tokens can cause abnormal responses from LLMs, potentially affecting their usage. However, we strictly adhere to ethical principles and do not condone any abuse or exploitation of these findings. Our research aims to raise awareness of these risks and contribute to a more secure LLM community. We have reported our findings to the respective LLM developers and are committed to collaborating with them to develop effective defenses and mitigation strategies. By working cooperatively, we promote responsible research practices and ensure the safe and beneficial use of LLMs.
\section{Background and Related Work}
\label{sec:background}

\subsection{Transformer-based LLMs}
\label{sec:background-transformer}

Self-attention~\cite{choromanski2020rethinking,luo2020simplified} is a core component of Transformer-based models, demonstrating strong modeling capabilities across various tasks. Given an input $X \in \mathbb{R}^{n \times d}$, where $n$ denotes the sequence length and $d$ denotes the dimension, self-attention linearly projects $X$ into query, key, and value representations, i.e., $Q$, $K$, and $V$. The attention scores matrix $A$ is then computed by taking the dot product between the query and key matrices, followed by a softmax normalization. The attention output is obtained by multiplying the attention scores with the value matrix.
\begin{equation}
A = \text{softmax}(\frac{QK^T}{\sqrt{d}})
\end{equation}
\begin{equation}
\text{Attention}(Q, K, V) = A \cdot V
\end{equation}
To analyze the behavior of Transformer-based models during sequence processing, we introduce the concept of \emph{attention patterns}, which can be extracted from the corresponding row $A[n]$ of the attention scores matrix $A$. In autoregressive generation tasks, the attention patterns only contain the attention weights between the current token and previously generated tokens.


The MLP module in Transformer-based models employs a gating mechanism similar to that of Gated Multi-Layer Perceptrons (gMLPs) \cite{liu2021pay}. Given an input $Y \in \mathbb{R}^{n \times d}$, where $n$ denotes the sequence length and $d$ denotes the dimension, the MLP first projects $Y$ to a higher-dimensional space using a linear transformation:
\begin{equation}
Z = YU
\end{equation}
where $U \in \mathbb{R}^{d \times d_m}$ is a learnable weight matrix. The transformed representation $Z$ is then split along the feature dimension into two matrices, $Z_1, Z_2 \in \mathbb{R}^{n \times d_m/2}$:
\begin{equation}
Z_1, Z_2 = \text{split}(Z)
\end{equation}
An activation function $\sigma$ is applied element-wise to $Z_1$ to obtain the \emph{MLP gate} $\sigma(Z_1)$. The MLP gate is then multiplied element-wise with the \emph{MLP data} $Z_2$ to produce the gated output $\tilde{Z}$:
\begin{equation}
\tilde{Z} = \sigma(Z_1) \odot Z_2 \
\end{equation}
Finally, another linear transformation is applied to map $\tilde{Z}$ back to the original dimension.
\begin{equation}
\text{Output} = \tilde{Z}W
\end{equation}
where $W \in \mathbb{R}^{d_m/2 \times d}$ is another learnable weight matrix.

The MLP gate $\sigma(Z_1)$ and MLP data $Z_2$ in the MLP can be seen as a special variant of the spatial gating unit in Transformer-based models. These two components work together to control the information flow and capture dependencies between tokens. By extracting and analyzing the MLP gate and MLP data, we can gain insights into how the model processes and responds to different types of input within the module.


\subsection{Glitch Token Phenomenon}

Tokenizer plays a key role in an LLM as it transforms a continuous text sequence into a list of discrete values called tokens~\cite{wang2024tokenization}. The tokens transformed from the training corpus form the vocabulary dictionary of LLMs, and the vocabulary dictionary in turn determines the capacity of LLMs to produce diverse and comprehensive output.
The rapid advancement of LLMs has brought attention to various anomalous phenomena~\cite{wang2024metmap,wang2024oopsla,li2024lockpicking,deng2024pandora,li2024digger,liu2023prompt,wang_ndss_2024}, one of which is the existence of ``glitch tokens''. These tokens exhibit anomalies in constructing the expected semantics, and are subsequently reflected in the abnormal and unexpected decoding in the LLM's output.


The glitch token phenomenon, first explored on the Lesswrong website, refers to anomalous tokens such as ``SolidGoldMagikarp'' and ``petertodd'' that cause unexpected and inaccurate results in language models like GPT-2 and GPT-J \cite{SolidGoldMagikarp2023, SolidGoldMagikarpII2023, SolidGoldMagikarpIII2023, petertodd_phenomenon2023}. Subsequent research examined the characteristics and instability of these tokens, revealing that even subtle changes in prompts can lead to significant differences and hallucinations in the generated results. The discovery of ``polysemous'' tokens, which produce different responses to repeated requests, further highlighted the prevalence and variability of the glitch token phenomenon in LLMs \cite{polysemous2023}.



Recently, Li et al.~\cite{li2024glitch} systematically investigated the glitch tokens with a proposed taxonomy covering their types and symptoms. They proposed three tasks, namely repetition, length and spelling, in their study to recognize glitch tokens. Their observation of the clustering distribution of glitch tokens in the word embedding spaces offers a novel perspective on the automatic identification of glitch tokens, making systematic detection feasible in LLMs containing billions, or even tens of billions of parameters.

The glitch token phenomenon uncovers the limitations and instability of LLMs when processing specific tokens. In this work, we aim to conduct a systematic and in-depth investigation of this phenomenon to gain a deeper understanding of the internal mechanisms of these models. This will provide valuable insights that can contribute to enhancing the robustness and reliability of LLMs in future applications.
\section{empirical study}

Our empirical study aims to explore an intuitive method for detecting glitch tokens. To this end, we investigate the differences in the model's behaviors when processing glitch tokens versus normal tokens. 
Two research questions are raised to guide the study:

\begin{itemize}[left=0pt]
    \item \textbf{RQ1 (Characteristics): What differences are exhibited between glitch tokens and normal tokens at the structural level of an LLM?} 
    \item \textbf{RQ2 (Ubiquity): Are the differences discovered in RQ1 prevalent in most LLMs?} 
\end{itemize}

To address the two RQs, we investigate the internal mechanisms of LLMs by analyzing the status of each layer in the transformer forward process of prediction. This involves examining the data flow of the intermediate layers as the model processes inputs. 

\subsection{Experiment Setup} 
To better understand the impact of glitch tokens on the model's internal output generation process, we conduct a series of experiments on the Llama-2-7b-chat model~\cite{touvron2023llama2}, shortly written as Llama2. Llama2 is a language model based on the Llama architecture. In our experiments, we set the temperature to 0 to eliminate randomness and ensure consistency in the model's responses. All other configurations are default.

We employ a unified approach to determine whether each token is glitchy or normal in this study based on existed definition of glitch tokens~\cite{2023_Glitch_Tokens}. While the symptoms of glitch tokens may vary across different tasks, we consistently utilize a repetition task to construct input sequences for glitch token identification.


In the context of our work, a \textbf{repetitive task} refers to a specifically designed experimental procedure to test the fidelity of a language model in reproducing input tokens. This task is utilized primarily for the identification and categorization of tokens based on their performance when repetitively prompted. Specifically, the repetitive task involves the following steps: 

\begin{enumerate}
    \item Formulating a prompt that requires the model to duplicate a specific token. The typical prompt structure is, ``Can you repeat the token `$\{token\}$' and return it back to me?'' This format is deliberately chosen to minimize contextual influence and focus purely on the token reproduction capability of the model.
    \item Submitting the prompt to the model, which is configured by setting the temperature parameter to zero. 
    \item Observing and analyzing the model's output to determine whether it accurately reproduces the input token. The output is deemed successful if the model returns the exact token as requested; otherwise, the token is classified as a glitch token.
\end{enumerate}

We traverse all the 32,000 tokens of the Llama2 model and eventually identify 6,425 glitch tokens from the entire vocabulary.To precisely capture intermediate layer outputs within the model, we resort to a transformer mechanistic interpretability tool named \emph{Transformer-lens}~\cite{TransformerLens}. Its hook technique enables real-time access of the activation values at all layers and allows code insertion into specific intermediate layers of the model. 
In this study, we insert hooks into all intermediate layers during the first forward of the tested model. This approach is chosen because the first forward comprehensively reflects the model's understanding of the input sequence and highlights the differences between normal and glitch tokens.

We select two key features to represent the model's internal output, i.e., \emph{attention patterns} and \emph{MLP status}. The attention patterns capture the relative importance and relationships between tokens, while MLP status is composed of MLP gate and MLP data~(Section~\ref{sec:background-transformer}), providing insights on how the model synthesizes and modulates new representations within the MLP module. 


\subsection{RQ1: Glitch Token Characteristics}\label{subsec:glitch_token_phenomenon}

\begin{figure}[t]
    \centering
    \includegraphics[width = 1\linewidth]{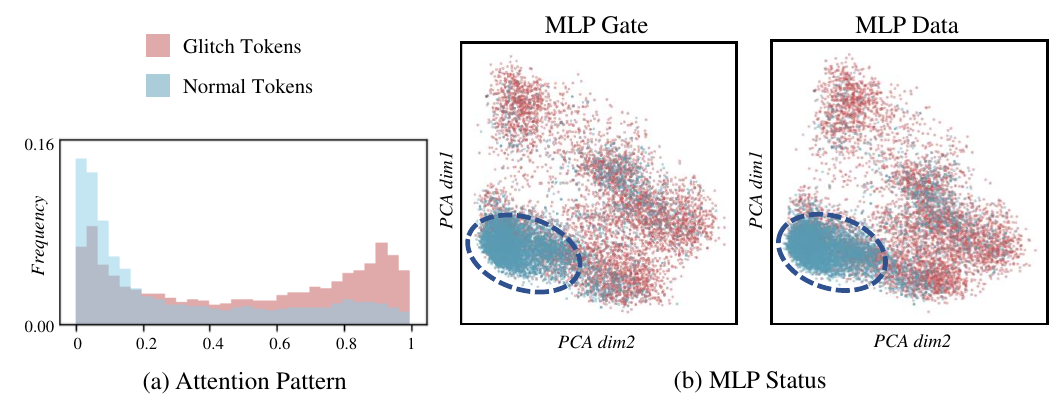}\vspace{-0.4cm}
    \caption{The distribution of attention patterns and MLP status for glitch tokens (shown in red color) and normal tokens (in blue color) in Llama2 model}\vspace{-0.4cm} 
    \label{fig:Llama2 distribution}
\end{figure}

We compare the extracted intermediate results of Llama2 model when processing prompts containing normal tokens and glitch tokens and observe significant disparity in attention patterns and MLP status.
The distribution of attention patterns for glitch tokens in some attention heads differs significantly from that of normal tokens.
To quantitatively analyze these differences, we randomly sample normal tokens in size of the same number as the pre-identified glitch tokens. 
To analyze the attention patterns, we create two sets of prompts: one containing 6,425 prompts with glitch tokens and another containing 6,425 prompts with normal tokens. We then compute the frequency distribution of attention patterns generated by these prompts, categorizing them into different value ranges. 
We visualize the results using a histogram, as shown in Figure~\ref{fig:Llama2 distribution} (a). The attention patterns of normal tokens (shown in blue color) generally cluster around lower ranges and exhibit a relatively smooth distribution. In contrast, the attention patterns for glitch tokens (in red color) display a comparably divergent and chaotic distribution.

\begin{figure*}[t]
    \centering
    \includegraphics[width = 0.9\linewidth]{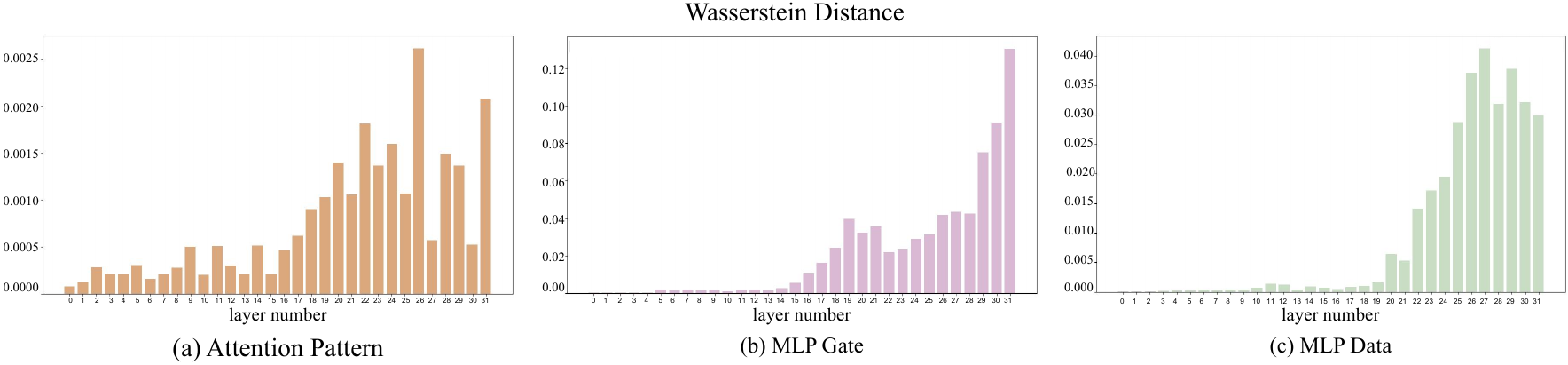}
    \vspace{-0.2cm}
    \caption{Wasserstein distance of the probability distributions between glitch tokens and normal tokens in different intermediate layers of Llama2 model}
    \vspace{-0.2cm}
    \label{fig:Wasserstein_Distance_across_layers}
\end{figure*}

Furthermore, the distribution characteristics of glitch tokens in the MLP status show deviations compared to normal tokens.
Due to the high dimension of MLP status values, we cannot visualize them in the same method used for attention patterns. 
Instead, we resort to PCA algorithm to convert the captured MLP status, i.e., MLP gate and MLP data, to two dimension values. 
We present the distribution of MLP gate and MLP data in scatter plots, as shown in Figure~\ref{fig:Llama2 distribution} (b).
It can be observed that both two representations of MLP status for normal tokens tend to cluster towards a centroid, forming a relatively dense and bounded distribution. In contrast, the MLP status of glitch tokens are highly dispersed and scattered.


\vspace{-0.2cm}
\findingbox{Finding 1}{Llama2 shows a significant disparity in attention patterns and MLP status when dealing with glitch tokens and normal tokens.}

To illustrate the anomalies across layers, we resort to the Wasserstein distance~\cite{vaserstein1969markov} to measure the magnitude of the differences in the intermediate layers outputs produced by normal and glitch tokens, and thereby reveal the distinctions in the model's internal mechanisms when processing these two groups of tokens. 
In this study, a larger Wasserstein distance indicates a greater distributional difference. 
Figure~\ref{fig:Wasserstein_Distance_across_layers} shows the Wasserstein distance in attention patterns and MLP status between normal and glitch tokens across different layers of the Llama2 model. 
We find that the differences caused by normal and glitch tokens per layer are not uniformly distributed. 
The attention patterns and MLP status exhibit greater differences in the downstream layers closer to the output, e.g., layers 19-31.
This finding suggests that the impact of glitch tokens, although may result in negligible erroneous results in front layers, is amplified along with the propagation, leading to unexpected outputs in the end.


\findingbox{Finding 2}{
The anomalous intermediate results caused by glitch tokens are not uniformly distributed across all layers of the model but are concentrated and amplified in specific key layers.
}

\subsection{RQ2: Ubiquity}



\begin{figure}[htbp]
    \centering
    \includegraphics[width = 1\linewidth]{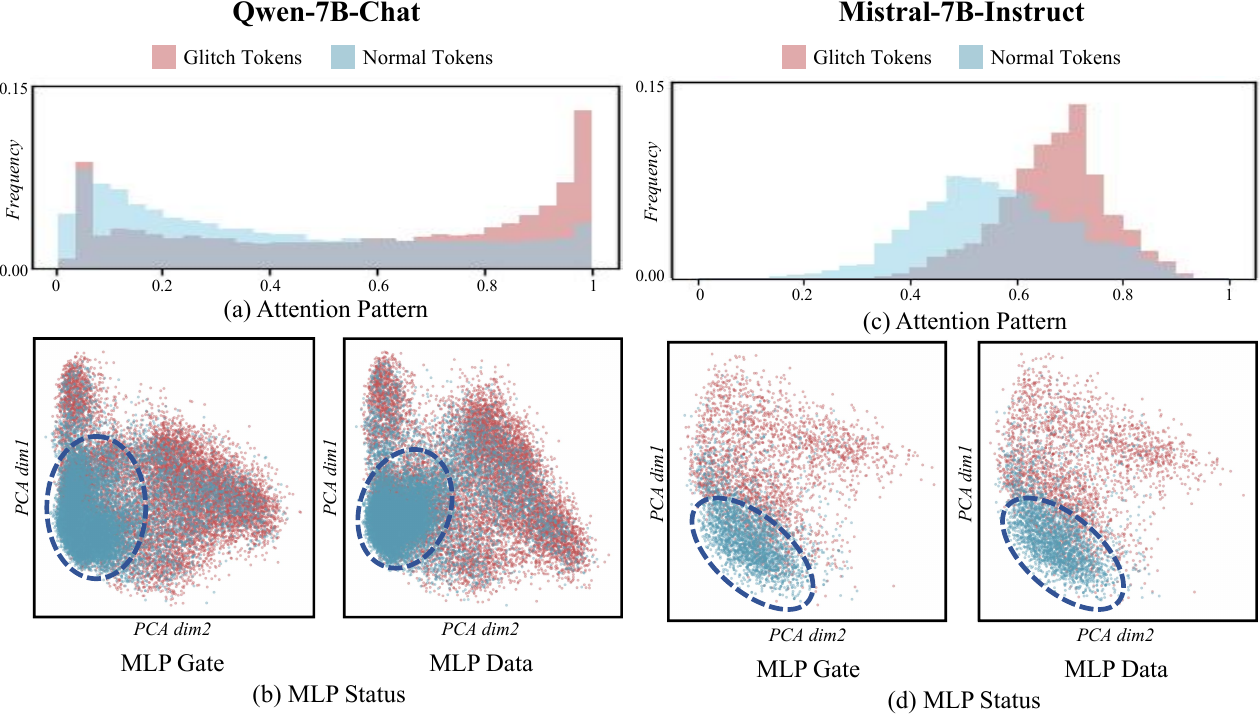}\vspace{-0.4cm}
    \caption{The example distribution of attention patterns, MLP gate and MLP data for glitch tokens (shown in red color) and normal tokens (in blue color) in \textbf{Qwen-7B-Chat} and \textbf{Mistral-7B-Instruct}.}\vspace{-0.4cm}
    \label{fig:Qwen distribution}
\end{figure}

In order to verify whether our previous findings exist in other LLMs, two additional LLMs, namely Qwen-7B-Chat model and Mistral-7B-Instruct model (shortly as Qwen and Mistral), are selected to complement our empirical study.  
As shown in Figure~\ref{fig:Qwen distribution}, the experimental results on these two models are similar to those of Llama2. 
The attention patterns of glitch tokens and normal tokens exhibit inconsistent distribution. For example, the attention patterns of normal tokens mainly fall within the range of $[0, 0.2]$ in the Qwen model, while the attention patterns of glitch tokens show a different shape and concentrates in the range of $[0.8, 1]$. 
Such inconsistency can also be observed in the Mistral model, evidenced by the attention values of normal tokens and glitch tokens primarily approximating around 0.5 and 0.7, respectively.
In terms of MLP status, the intermediate layers tend to produce outputs in a centroid-based cluster for normal tokens. However, the MLP status values of glitch tokens are overall chaotic and disseminated.

\findingbox{Finding 3}{
We identify similar differences exhibited between normal and glitch tokens at the intermediate layers of different LLMs.
}

The findings of RQ1 provide insights for the subsequent detection and fix of glitch tokens, while the finding in RQ2 offers factual evidence for the broad application of our approach in LLMs.

\section{Methodology}

\begin{figure*}[htbp]
\centering
\includegraphics[width=0.9\textwidth]{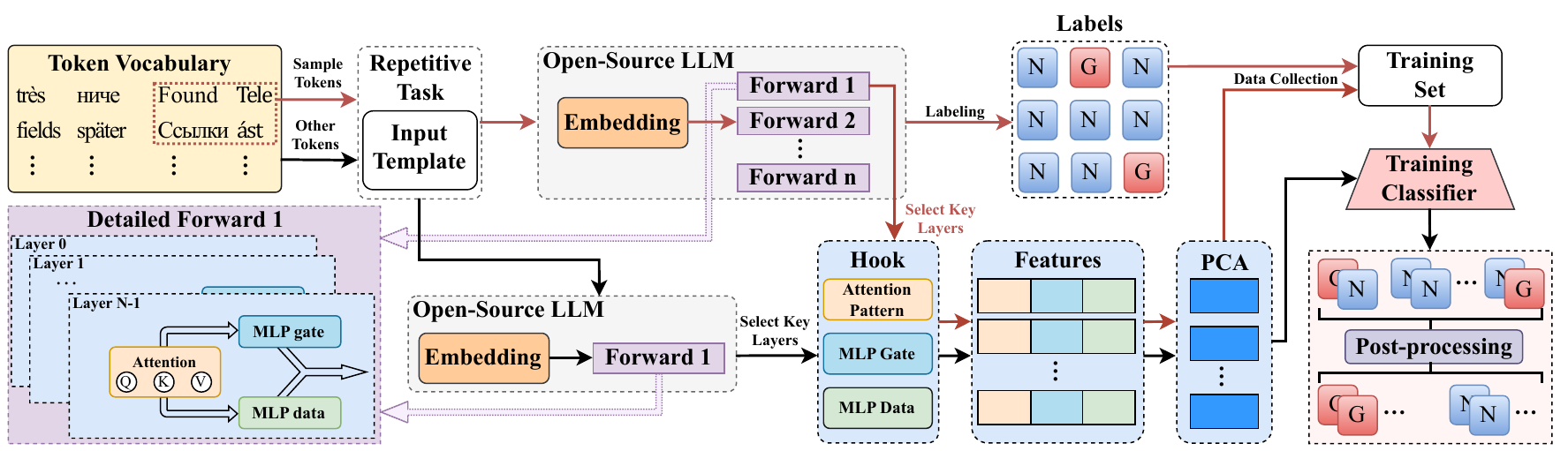}
\vspace{-0.4cm}
\caption{\tool workflow for detecting glitch tokens. The red arrows represent the data flow during the training process, while the black arrows represent the data flow during the detection process.
}
\vspace{-0.2cm}
\label{fig: detection workflow}
\end{figure*}

Based on the findings from our empirical study, we propose the \tool algorithm, which aims to achieve automatic detection and fix of glitch tokens by analyzing the internal activation states of LLMS. Our approach consists of two main ideas:

\begin{enumerate}[left=0pt]
\item Leveraging the differences in model activation values when processing glitch tokens and normal tokens to achieve rapid screening of glitch tokens. By designing an anomaly detection algorithm, we can identify and label potential glitch tokens based on their activation features extracted from specific layers, which we refer to as key layers~(detailed in Section~\ref{Kay layers introduction}). These key layer features are crucial for detecting glitch tokens.
\item Fixing errors caused by glitch tokens by adjusting the model's intermediate results. We designed a series of experiments where we automatically adjusted the output values of the model's intermediate layers and observed the impact on the final output. 
\end{enumerate}

As elucidated in Section~\ref{subsec:glitch_token_phenomenon}, glitch tokens in the model predominantly affect the downstream layers close to the output, notably impacting attention patterns and MLP status in specific key layers. This revelation is instrumental in shaping our approach of \tool. By strategically focusing our efforts on key layers, we can significantly reduce computational overhead while achieving a comparable level of detection and fix efficacy to traversing all layers. Thus, this approach not only optimizes the efficiency of our methods but also ensures that our interventions are targeted where they are most needed. For a detailed introduction to the selection strategy of key layers, please refer to Section~\ref{Kay layers introduction}.

Based on these ideas, we designed \tool, which integrates detection and fix algorithms. \tool's detection algorithm identifies and locates glitch tokens causing model output errors by analyzing intermediate layer activation states. It randomly samples a subset of tokens from the vocabulary, tests them using repetitive tasks, and extracts activation features from key layers. These features undergo dimensionality reduction and are labeled based on repetitive task outcomes. A classifier is then trained with the labeled data to assess unknown tokens. Finally, the remaining tokens are detected individually using the trained SVM classifier, and predictions are verified through repetitive tasks.

\tool's fix algorithm corrects the anomalous activation patterns of glitch tokens by adjusting activation values in the model's intermediate layers, eliminating their negative impact on the model output. It first calculates activation statistics of normal tokens in key layers and identifies neurons that are consistently activated or silent in most normal tokens. Then, it compares these neurons' activations between normal and glitch tokens, calculating suppression ratio coefficients for anomalous activations and activation values to promote silent neurons. Finally, it rectifies the activation values of glitch tokens in key layers based on these coefficients and values, automatically fixing the glitch tokens.

\vspace{-0.3cm}
\subsection{Detecting Glitch Tokens via \tool}

\tool identifies and detects glitch tokens that cause model output errors by analyzing the intermediate layer activation states of Transformer language models when processing tokens. The main workflow of the algorithm is shown in Figure \ref{fig: detection workflow}. The detection algorithm of \tool consists of three main steps: feature extraction and dimension reduction, SVM-based glitch token classifier, and glitch token identification and validation.

\subsubsection{\textbf{Feature Extraction and Dimension Reduction}}
\tool adopts a random sampling strategy to select samples from the model's token vocabulary $V$ to form the sample set $S$. The choice of sampling rate $\gamma$ needs to balance between sample size and computational efficiency. A larger $\gamma$ leads to a larger sample size and more accurate detection results but also incurs higher computational costs. Conversely, a smaller $\gamma$ results in a smaller sample size and faster computation but may affect the detection performance. Through experiments, we determined that when $\gamma$ is in the range of [0.1, 0.3], \tool achieves a good balance between detection performance and efficiency.

\tool uses a unified repetitive task to construct input sequences for glitch token identification in the sample set $S$. For tokens in the sample set $S$, the algorithm assigns corresponding category labels according to the output results of the repetitive task. At the same time, we extract the attention pattern, MLP gate and MLP data features in the model's first forward process. However, the dimension of the original feature tensors is high, and directly using them to train the classifier would lead to excessive computational costs. To improve computational efficiency, we adopt a dimension reduction strategy. The PCA algorithm~\cite{abdi2010PCA} is applied to reduce the dimension, mapping the original high-dimensional features to a low-dimensional subspace while maximally preserving the discriminative information of the features. Through experiments, we found that when the dimension $P$ of the reduced features is in the range of [50, 200], a good balance between computational efficiency and information retention can be achieved. Therefore, we set $P=75$ as the default dimension reduction parameter.

\subsubsection{\textbf{SVM-based Glitch Token Classifier}}

\tool trains an SVM classifier using the low-dimensional feature representation matrix $F$ of the sample set $S$ and the corresponding category labels. The trained SVM classifier will be used for subsequent glitch assessment of unknown tokens. SVM is a binary classification algorithm that is particularly suitable for problems with high-dimensional feature spaces, which aligns well with the characteristics of the glitch token detection task in \tool. Besides, SVM has higher computational efficiency and shorter training time when handling high-dimensional features compared to other binary classification algorithms. That helps achieve rapid real-time detection of glitch tokens in \tool. Therefore, we adopt SVM as the classifier in the detection algorithm of \tool.

\subsubsection{\textbf{Glitch Token Identification and Validation}}

The tokens in the token vocabulary that were not sampled are individually detected. The token to be detected is input into the same repetitive task as in the training phase, and its features attention patterns, MLP gate, and MLP data are extracted in the model's first forward module. Subsequently, the trained SVM classifier is used to make predictions based on the extracted features. 

If a token is predicted as ``glitchy'', the algorithm will input this token into the model and further use the repetitive task to validate it. If the model can correctly repeat the token, we consider it as a potential normal token, and the SVM classifier may have made a false positive prediction.  Through this post-processing step, \tool can effectively reduce the false positive rate of glitch token detection and improve the precision of detection. Finally, \tool outputs two sets: the set of glitch tokens $G$ and the set of normal tokens $N$.

\begin{algorithm}[htbp]
\small
\SetAlgoLined
\caption{\tool (Detection)}
\label{Algo:Detection}
\KwIn{Token Vocabulary $V$; PCA Dimension $P$; Sample Rate $\gamma$; KeyLayers[]}
\KwOut{Glitch token set $G$; Normal token set $N$}
$S \gets \text{randomSample}(V, \gamma)$; \label{algo:detection:Sample subset}

\ForEach{$ \text{Token} \in S$}{ \label{algo:detection:Feature extraction}
    $ \text{SampleFeatures} \gets \text{hookModel}(\text{Token}, \text{KeyLayers})$\;
    $ \text{SampleLabels} \gets \text{validateGlitch}(\text{Token})$\;
}\label{algo:detection:Labeling}
$F \gets \text{PCA}(\text{SampleFeatures}, P)$\; \label{algo:detection:Feature PCA}
$ \text{Classifier} \gets \text{trainClassifier}(F, \text{SampleLabels})$\; \label{algo:detection:Classifier Training}

\ForEach{$\text{Token} \in V$}{\label{algo:detection:Predict Start}
    \If{$\text{Token} \notin S$}{
        $\text{Feature} \gets \text{PCA}((\text{hookModel}(\text{Token}, \text{KeyLayers})), P)$\; 
        
        \eIf{$\text{classify}(\text{Classifier}, \text{Feature}) == \text{`Normal'}$}{
            $N \gets \text{Token}$\; \label{algo:detection:Predict End}
        }
        {
            \eIf{$\text{validateGlitch}(\text{Token}) == \text{`Glitch'}$}{
                \label{algo:detection:Post Process Start}
                $G \gets \text{Token}$\;
            }
            {
                $N \gets \text{Token}$;
            }\label{algo:detection:Post Process End}
        }
    }
}
\end{algorithm}
\vspace{-0.4cm}
\subsubsection{\textbf{Detection Process Algorithm}}
The pseudocode for the detection algorithm of \tool is shown in Algorithm~\ref{Algo:Detection}. Initially, the algorithm samples a subset of tokens ($S$) from the vocabulary ($V$) based on a predefined sampling rate ($\gamma$) (line~\ref{algo:detection:Sample subset}). For each token in this subset, the algorithm extracts relevant features using a transformer model over specified key layers and assigns labels indicating whether each token is glitchy (lines~\ref{algo:detection:Feature extraction}-\ref{algo:detection:Labeling}). 

Following the feature extraction, the algorithm applies PCA to reduce the dimension of these features to a lower-dimensional space ($P$) (line~\ref{algo:detection:Feature PCA}). The reduced feature set ($F$) is then used to train a SVM classifier with the assigned glitch labels (line~\ref{algo:detection:Classifier Training}).

For tokens not included in the initial sample, the algorithm uses the trained SVM classifier to predict whether each token is normal or a glitch (lines~\ref{algo:detection:Predict Start}-\ref{algo:detection:Predict End}). Tokens classified as glitches will undergo a further validation step to confirm their status, effectively minimizing the false positive rate (lines~\ref{algo:detection:Post Process Start}-\ref{algo:detection:Post Process End}). Finally, the algorithm compiles and outputs two sets, namely glitch token set ($G$) and normal token set ($N$).

\vspace{-0.2cm}
\subsection{Fixing Glitch Tokens via \tool}

\begin{figure}[t]
\centering
\includegraphics[width=\linewidth]{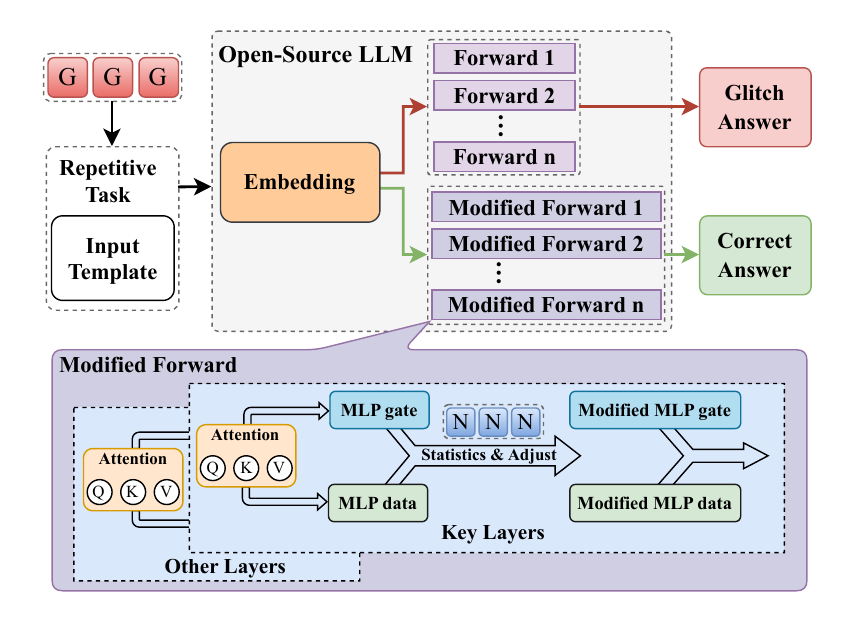}\vspace{-0.4cm}
\caption{\tool workflow for fixing glitch tokens.}\vspace{-0.4cm}
\label{fig: fix workflow}
\end{figure}

We further explore the possibility of correcting the anomalous activation patterns of glitch tokens to normal patterns by adjusting the activation values of the model's intermediate layers, thereby eliminating their negative impact on the model output. 
Based on the idea of adjusting neuron activation values to eliminate the influence of glitch tokens, we focus on two types of neurons in normal tokens, i.e., the neurons that are activated in the vast majority of normal tokens, and the neurons that are not activated in any normal tokens. 
Then we compare the differences in activation values of these key neurons between normal tokens and glitch tokens. By simulating those normal activation patterns, we achieve adaptively adjustment of the activation values of glitch tokens. The overall approach is illustrated in Figure~\ref{fig: fix workflow}.

\subsubsection{\textbf{Normal Token Activation Value Statistics}}

\tool first randomly samples a subset of normal tokens, i.e., $N' \subset N$, to calculate the activation value distribution of normal tokens in the target layers. We set the sampling rate to $\gamma$, which is consistent with the sampling rate in the glitch token detection phase.

For the MLP module in each layer, we calculate the activation statistics of the tokens in the normal token set $N'$. We define two sets of neuron indices as $Neun^\uparrow$ and $Neun^\downarrow$ based on their activation patterns, where $Act[i]$ represents the activation value of the $i$-th neuron in the MLP module, and $m$ is the predefined threshold.

\begin{align}
Neun^\uparrow &= \{i \,| {Act}[i] > m \text{ for over 99\% of tokens in } N'\} \\
Neun^\downarrow &= \{i \,| {Act}[i] \leq m \text{ for all tokens in } N'\}
\end{align}

$Neun^\uparrow$ denotes the set of key neurons that exhibit high activation levels, surpassing predefined threshold $m$, across sample token set $N'$. Given that activated neurons constitute a small proportion of the total neurons, we consider a neuron to be a key neuron if it is activated in over 99\% of the tokens. Conversely, $Neun^\downarrow$ represents the set of key neurons that exhibit consistently low activation levels, falling below $m$. We consider a neuron to be a key neuron in $Neun^\downarrow$ if its activation level remains below the threshold $m$ for all tokens in $N'$. These neurons can be considered as the key features for suppressing noise, as they are consistently inactive for normal tokens. By identifying these two sets of neuron indices based on their activation patterns, we can create a profile of the expected behavior of normal tokens at each MLP module. Those methods ensure the recorded neurons take the most informative features when we mitigate the influence of noise and irrelevant information caused by glitch tokens.

Note that we only adjust the activation values of the MLP module and not the attention patterns. Attention patterns capture the relative importance between tokens, and modifying them may disrupt token dependencies and introduce noise. In contrast, MLP activation values reflect the model's understanding of each token independently, and adjusting these values has less impact on token relationships. By only adjusting MLP activation values, we aim to fix glitch tokens without introducing additional noise.

\subsubsection{\textbf{Activation Value Adjustment}}
\label{subsubsec:adjust}
When the hooked model processes detected glitch tokens, \tool intervenes in this process, making trend-based adjustments to the neurons identified by $Neun^\uparrow$ and $Neun^\downarrow$. For neurons in $Neun^\uparrow$ that should be activated but have insufficient activation in glitch tokens, the algorithm uses $\beta$ as an amplification factor to promote their activation. Conversely, for neurons in $Neun^\downarrow$ that should be suppressed but are abnormally activated in glitch tokens, the algorithm uses $\alpha$ as a reduction factor to suppress their anomalous activation. The factors $\beta$ and $\alpha$ are named adjustment factors, which play a crucial role in the glitch token fixing process. For the calculation of $\beta$ and $\alpha$, we first calculate the average activation value difference ${\Delta Act}^{\uparrow}$ of glitch tokens relative to normal tokens on highly activated neurons ($Neun^\uparrow$), and the average activation value ratio ${\Delta Act}^{\downarrow}$ on lowly activated neurons ($Neun^\downarrow$). 


\vspace{-0.3cm}
\begin{align}
{\Delta Act}^\uparrow &= \frac{1}{|Neun^\uparrow|} \sum_{i \in Neun^\uparrow} (Act_{\text{normal}}[i] - Act_{\text{glitch}}[i]) \\
{\Delta Act}^\downarrow &= \frac{1}{|Neun^\downarrow|} \sum_{i \in Neun^\downarrow} \left(\frac{Act_{\text{glitch}}[i]}{Act_{\text{normal}}[i]}\right)
\end{align}

Subsequently, through linear transformation and range restriction, the algorithm maps ${\Delta Act}^{\uparrow}$ and ${\Delta Act}^{\downarrow}$ to appropriate numerical intervals to obtain the values of $\beta$ and $\alpha$, respectively. 
\vspace{-0.1cm}
\begin{equation}
\beta = k_1 \cdot {\Delta Act}^\uparrow + b_1
\end{equation}
\vspace{-0.4cm}
\begin{equation}
\alpha = k_2 \cdot {\Delta Act}^\downarrow + b_2
\end{equation}

The constants $k_1$, $b_1$, $k_2$, and $b_2$ are derived through an adaptive process tailored to the specific dynamics of each model. A set of default values is provided, which can be adjusted based on empirical data to optimize the correction process for different types of models. They are crucial for ensuring $\beta$ and $\alpha$ effectively modulate neuron activations while maintaining system stability and performance.

After the adjustment of the MLP  activation values is completed, we input the corrected activation values back into the subsequent layers, allowing the model to continue the forward propagation until the final fixed result is output. This process corrects on each token in the detected glitch token set $G$.




\begin{algorithm}[htbp]
\small
\SetAlgoLined
\caption{\tool (Fix)}
\label{Algo:Fix}
\KwIn{Glitch token set $G$; Normal token set $N$; Sample Rate $\gamma$; Threshold $m$; KeyLayers[]}
$N' \gets \text{randomSample}(N, \gamma)$\; \label{algo:fix:Sample}
$Neun^\uparrow, Neun^\downarrow \gets \text{statisticsNeuron}(N', m)$\; \label{algo:fix:Statistics key neurons}
$\beta \gets \text{statisticsBeta}(N', Neun^\uparrow)$\; \label{algo:fix:Statistics beta}
$\alpha \gets \text{statisticsAlpha}(N', Neun^\downarrow)$\; \label{algo:fix:Statistics alpha}
\ForEach{$\text{Token} \in G$}{\label{algo:fix:Adjustment start}
    \For{$\text{Layer} \in \text{KeyLayers}$}{
        $Activation \gets \text{hookModel}(\text{Token}, \text{Layer})$\; \label{algo:fix:Hook activation start}
        \ForEach{$\text{Neuron} \in Neun^\uparrow$}{ \label{algo:fix:Adjustment beta start} 
            $Act[\text{Neuron}] \gets Act[\text{Neuron}] + \beta$\;
        }\label{algo:fix:Adjustment beta end} 
        \ForEach{$\text{Neuron} \in Neun^\downarrow$}{ \label{algo:fix:Adjustment alpha start}
            $Act[\text{Neuron}] \gets Act[\text{Neuron}] / \alpha$\;
        }\label{algo:fix:Adjustment alpha end}
        
        $\text{hookModel}(\text{Token}, \text{Layer}) \gets Activation$; \label{algo:fix:Hook activation end}
    }
}\label{algo:fix:Adjustment end}
\end{algorithm}
\vspace{-0.5cm}

\subsubsection{\textbf{Fixing Process Algorithm}}
Based on Section~\ref{subsubsec:adjust}, we present the pseudocode for the fix algorithm of \tool in Algorithm~\ref{Algo:Fix}. The algorithm begins by sampling a subset ($N'$) of normal tokens from the full set of normal tokens ($N$) (line~\ref{algo:fix:Sample}). Then it computes the statistical distribution of activation values across key neurons in the subset, distinguishing $Neun^\uparrow$ and $Neun^\downarrow$ (line~\ref{algo:fix:Statistics key neurons}). After that the algorithm calculates the adjustment factors $\beta$ and $\alpha$ for the identified key neurons (line~\ref{algo:fix:Statistics beta}-\ref{algo:fix:Statistics alpha}).

For each glitch token in the set $G$, the algorithm iteratively applies these adjustments across specified layers (i.e., KeyLayers) of the model (lines~\ref{algo:fix:Adjustment start}-\ref{algo:fix:Hook activation start}). Neurons in $Neun^\uparrow$ have their activation values increased by $\beta$, amplifying their response to mimic normal activation patterns (lines~\ref{algo:fix:Adjustment beta start}-\ref{algo:fix:Adjustment beta end}). Conversely, neurons in $Neun^\downarrow$ have their activation values reduced by dividing by $\alpha$, suppressing any abnormal activations (lines~\ref{algo:fix:Adjustment alpha start}-\ref{algo:fix:Adjustment alpha end}). Each adjusted activation is reintegrated into the model's processing flow, allowing it to continue with forward propagation with the corrected values (line~\ref{algo:fix:Hook activation end}).

\subsection{Key Layers Selection}\label{Kay layers introduction}
In the design of our \tool's detection and fix algorithms, we focused on exploiting the attention pattern and MLP status features within certain \emph{key layers}. The rationale behind selecting these key layers stems primarily from our empirical findings outlined in Finding 2~(Section~\ref{subsec:glitch_token_phenomenon}), which highlighted that glitch tokens predominantly affect the model's downstream layers closer to the output. For instance, in Llama2, this pertains to layers 19 to 31. Further refining our selection, we encountered a counter-intuitive discovery: modifying features in layers exceedingly close to the output paradoxically diminished the effectiveness of our fix algorithms.

The layers preceding the final output are crucial for tailoring responses based on preceding computations; alterations in these layers can disrupt representational balances, leading to degraded performance. In our approach for Llama2, we designated layers 19 to 28 as key layers, optimally positioned in the middle to lower sections of the model's architecture. This strategic placement ensures that our interventions effectively mitigate the effects of glitch tokens while preserving the model's robustness.

\section{evaluation of Glitch Token detection}

\subsection{Experiment setup}
\noindent\textbf{Experiment Environment.}
All experiments are performed on a workstation with Ubuntu 22.04.3 LTS and 250GB memory, and 2 A100 GPU with 80GB memory each. 

\noindent\textbf{LLM Selection.} We thoroughly evaluated our proposed method using a diverse set of computational models. We selected five widely recognized, open-source models, including Llama-2-7b-chat~\cite{touvron2023llama2}, Mistral-7B-Instruct-v0.1~\cite{jiang2023mistral}, Qwen-7B-Chat~\cite{bai2023qwen}, Gemma-2b-it~\cite{gemma_2024}, and Yi-6B-Chat~\cite{ai2024yi}. These models served as the subjects for our in-depth analysis, allowing us to assess the versatility and effectiveness of our method across various real-world applications. Table \ref{model_information} provides an overview of these models' parameters. 

\noindent\textbf{Evaluation of Detection Baselines.} To evaluate the performance of \tool, we compared it with two implemented benchmark schemes and a recent testing method, \textsc{GlitchHunter}. 
\begin{enumerate}[left=0pt]
    \item \textbf{Exhaustive Search:} Each token in the token list is individually fed into the model, which performs tasks such as paraphrasing, spelling, and length calculation for each token.
    \item \textbf{Rule-based Random Sampling:} First, randomly select half of the tokens from the language model to form a candidate set. Since common English words typically do not become glitch tokens, use the Natural Language Toolkit (NLTK) to remove high-frequency English words from the candidate set. The remaining tokens are considered potential glitch tokens.
    \item \textbf{\textsc{GlitchHunter}:} This is the state-of-the-art automated detection method for glitch tokens~\cite{li2024glitch}. 
\end{enumerate}

\noindent\textbf{Evaluation Metrics of Detection.} For efficiency evaluation, we consider the \textbf{Time Cost} required to process all glitch tokens in the complete token list of a model. For \tool, it encompasses the total duration including feature extraction, classifier training, identification and validation. For effectiveness evaluation, we consider \textbf{True Positive}, \textbf{Precision}, \textbf{Recall} and \textbf{F1-Score}.

\noindent\textbf{Evaluation Settings.}
In our detection experiments, we evaluated the performance of \tool with SVM regularization parameter and degree\cite{vapnik1995nature,scholkopf2002learning} set to $C=1, degree=3$. 
We employ $\gamma=0.1$ for random sampling and principal components to $P=75$ for PCA. 
For the rule-based random sampling methods, we conducted 100 independent experiments and averaged the results to obtain statistically significant conclusions. For the \textsc{GlitchHunter} method, we used the default settings from the original paper~\cite{li2024glitch}. 


\subsection{RQ3 (Efficient Detection): How efficient is our approach in identifying glitch tokens across different LLMs?}

To evaluate the efficiency of GlitchProber, time overhead and the accuracy comparison results of various methods on five large models are shown in Table \ref{detection timecost result} and Table \ref{detection performance results}.



Our experimental results demonstrate that \tool reached a detection efficiency advantage. 
Meanwhile, \tool achieves a 100\% precision which matches the performance of both \textsc{GlitchHunter} and the exhaustive search benchmark method. This signifies \tool's minimal false positive rate. Furthermore, \tool achieves a recall rate of 64.47\%, surpassing \textsc{GlitchHunter}'s 26.52\%. The F1-score indicates that \tool strikes a fine balance between precision and recall, efficiently detecting glitch tokens while maintaining high accuracy. 

\begin{tcolorbox}[boxsep=3pt,left=4pt,right=4pt,top=2pt,bottom=2pt,colback=gray!20,colframe=gray!80!black,title=Answer to RQ3]
    \tool achieves perfect accuracy across all test cases with low time overhead, exhibiting superior stability and performance in glitch token detection. Its efficiency gains are primarily attributed to its strategic adoption of small-scale sampling and intermediate layer feature extraction techniques, significantly enhancing detection efficacy.
\end{tcolorbox}

\begin{table}[t]
\centering
\caption{Summary of models in evaluation \label{model_information}}\vspace{-0.3cm}
\setlength{\tabcolsep}{2pt}
\scriptsize 
\resizebox{\linewidth}{!}{
    \begin{tabular}{l|c c c c c}
    \hline
    \textbf{Model Name} & \textbf{\makecell{Number of\\Parameters}} & \textbf{\makecell{Vocabulary\\Size}} & \textbf{\makecell{Hidden\\Layers}} & \textbf{\makecell{Intermediate\\Size}} & \textbf{\makecell{Attention \\ Heads}} \\ \hline
    Llama-2-7b-chat & 6.74B & 32,000 & 32 & 11,008 & 32 \\ \hline
    Mistral-7B-Instruct-v0.1 & 7.24B & 32,000 & 32 & 14,336 & 32 \\ \hline
    Qwen-7B-Chat & 7.72B & 151,936 & 32 & 22,016 & 32 \\ \hline
    Gemma-2b-it & 2.51B & 256,000 & 18 & 16,384 & 8  \\ \hline
    Yi-6B-Chat & 6.06B & 64,000 & 32 & 11,008 & 32 \\ \hline
    \end{tabular}
}\vspace{-0.1cm}
\end{table}

\begin{table}[t]
\centering
\setlength{\tabcolsep}{1.5pt}
\caption{Time cost comparison of \tool and other baselines on different LLMs.}\vspace{-0.4cm}
    \resizebox{\linewidth}{!}{
        \begin{tabular}{l|c|c|c}
        \hline
        \textbf{Test Model} & \textbf{Exhaustive Search} & \textbf{\textsc{GlitchHunter}} & \textbf{\tool  (ours)} \\ \hline
        Llama-2-7b-chat & 619min 43s & 74min 11s & \textbf{61min 38s} \\
        Mistral-7B-Instruct-v0.1 & 651min 17s & 64min 26s & \textbf{42min 39s} \\
        Qwen-7B-Chat & 2,228min 23s & 720min 42s & \textbf{92min 48s} \\
        Gemma-2b-it & 3,575min 9s & 681min 16s  & \textbf{96min 43s} \\
        Yi-6B-Chat & 974min 4s & 825min 25s & \textbf{140min 57s} \\
        \hline
        Average Time Cost & 1,609min 42s & 473min 11s & \textbf{89min 9s} \\ \hline
        \end{tabular}
    }
    \label{detection timecost result}
\end{table}

\begin{table}[t]
    \centering
    \caption{Performance comparison of \tool and other baselines on different LLMs}\vspace{-0.3cm}
    \setlength{\tabcolsep}{2pt}
    \renewcommand\arraystretch{0.9}
    \resizebox{\linewidth}{!}{
    \begin{tabular}{l|c|c|c|c}
        \hline
        \textbf{Test Model} & \textbf{Metric} & \textbf{\makecell[ct]{Rule-based\\Random Sampling}} & \textbf{\textsc{GlitchHunter}} & \textbf{\tool} \\ \hline
        \multirow{4}{*}{Llama-2-7b-chat} & TP & 2,936 & 1,955 & 4,446 \\
        & Precision & 24.74\%   & 100.00\%  & 100.00\% \\
        & Recall    & 45.70\%   & 30.43\%   & 69.22\% \\
        & F1-score  & 0.3210    & 0.4724    & 0.8181 \\ \hline
        \multirow{4}{*}{Mistral-7B-Instruct-v0.1} & TP & 1,288 & 1,233 & 1,873 \\
        & Precision & 11.44\%   & 100.00\%  & 100.00\% \\
        & Recall    & 46.35\%   & 44.37\%   & 67.41\% \\
        & F1-score  & 0.1836    & 0.6147    & 0.8053 \\ \hline
        \multirow{4}{*}{Qwen-7B-Chat} & TP & 15,419 & 4,031 & 19,366 \\
        & Precision & 21.04\%   & 100.00\%  & 100.00\% \\
        & Recall    & 50.24\%   & 14.42\%   & 63.08\% \\
        & F1-score  & 0.2966    & 0.2521    & 0.7736 \\ \hline
        \multirow{4}{*}{Gemma-2b-it} & TP & 13,777 & 3,240 & 17,387 \\
        & Precision & 11.30\%   & 100.00\%  & 100.00\% \\
        & Recall    & 49.27\%   & 10.56\%   & 62.18\% \\
        & F1-score  & 0.1838    & 0.1910    & 0.7668 \\ \hline
        \multirow{4}{*}{Yi-6B-Chat} & TP & 3,215 & 2,662 & 4,900 \\
        & Precision & 13.10\%   & 100.00\%  & 100.00\% \\
        & Recall    & 39.67\%   & 32.84\%   & 60.45\% \\
        & F1-score  & 0.1969    & 0.4944    & 0.7535 \\ \hline
        \multirow{3}{*}{Average Performance} & Precision & 16.32\% & 100.00\% & 100.00\% \\
        & Recall & 46.24\% & 26.52\% & 64.47\% \\
        & F1-score & 0.2364 & 0.4049 & 0.7835 \\ \hline
        \end{tabular}
  }
    \label{detection performance results}
\end{table}
\vspace{-0.4cm}
\subsection{RQ4 (Ablation Study): How do the different components of \tool affect the detection results?}


To assess the importance of the components in \tool, we performed an ablation study across five models. We developed two variants: \toolnopca and \toolnopost. \toolnopca omits the PCA during feature processing, while \toolnopost eliminates the final token validation steps in the original \tool. The comprehensive results are shown in Table~\ref{tab:ablation_time_mem}.

\begin{table}[t]
    \centering
\setlength{\tabcolsep}{1.5pt}
    \caption{Comparison of performance and memory usage for \tool with different feature configurations across various language models.}\vspace{-0.4cm}
    \label{tab:ablation_time_mem}
    \resizebox{\linewidth}{!}{
    \begin{threeparttable}
        \begin{tabular}{l|c|c|c|c}
        \hline
        \textbf{Model} & \textbf{Metrics} & \textbf{\tool} & \textbf{\toolnopost} & \textbf{\toolnopca} \\
        \hline
        \multirow{2}{*}{Llama-2-7b-chat} & F1-Score & 0.8529 & 0.6097 & -- \\
        & Memory & 103.71GB & 101.22GB & 250.00GB \\
        \hline
        \multirow{2}{*}{Mistral-7B-Instruct-v0.1} & F1-Score & 0.8652 & 0.5429 & --  \\
        & Memory & 107.11GB & 102.67GB & 250.00GB \\
        \hline
        \multirow{2}{*}{Qwen-7B-Chat} & F1-Score & 0.8854 & 0.5510 & --  \\
        & Memory & 109.03GB & 101.20GB & 250.00GB \\
        \hline
        \multirow{2}{*}{gemma-2b-it} & F1-Score & 0.8143 & 0.4226 & --  \\
        & Memory & 131.04GB & 127.96GB & 250.00GB \\
        \hline
        \multirow{2}{*}{Yi-6B-Chat} & F1-Score & 0.8718 & 0.4507 & --  \\
        & Memory & 83.32GB & 82.76GB & 250.00GB \\
        \hline
        \end{tabular}
        \begin{tablenotes}
      \item Note: `--' denotes incomplete experiment due to exceeding the maximum memory of our server of 250.00GB. 
    \end{tablenotes}
  \end{threeparttable}
    }\vspace{-0.6cm}
\end{table}

Table~\ref{tab:ablation_time_mem} offers a detailed comparative analysis of \tool alongside its variants. With respect to the F1-score, \tool markedly surpasses \toolnopost, demonstrating average improvements of 0.32. These findings underscore the vital importance of implementing robust post-classification enhancements to boost overall performance. Furthermore, the absence of PCA in \toolnopca leads to substantial increases in memory usage beyond the maximum capacity of our server, resulting in the incompletion of \toolnopca. This starkly highlights the necessity of dimensionality reduction as a means to optimize resource allocation and ensure system stability.

\begin{tcolorbox}[boxsep=3pt,left=4pt,right=4pt,top=2pt,bottom=2pt,colback=gray!20,colframe=gray!80!black,title=Answer to RQ4]
The PCA dimensionality reduction and the post-process are crucial components for \tool. Omitting either of these components leads to a significant decrease in effectiveness, undermining the utility of \tool.
\end{tcolorbox}

\section{evaluation of glitch token fix}

\subsection{Experiment setup}

\textbf{Evaluation Baselines of Fix.} Due to the lack of existing methods for fixing glitch tokens, we compared \tool with a benchmark scheme: a rule-based fix method, to verify its effectiveness. Specifically, the rule-based fix method does not rely on specific activation value differences to determine $\alpha$ and $\beta$. Instead, it directly uses a fixed value to adjust activation values at the same neuron positions as \tool.

\noindent\textbf{Evaluation Metrics of Fix Experiments.} We used two test metrics to evaluate the performance of the fix methods: the number of repaired glitch tokens (Repaired Tokens) and the repair rate (Repair Rate). The repair rate represents the proportion of glitch tokens successfully repaired out of all glitch tokens. It is calculated using the following formula: 
\vspace{-0.2cm}
\begin{equation}
    \text{Repair Rate}=\frac{\text{Repaired Token Number}}{\text{Total Glitch Token Number}}
\end{equation}
These two metrics can intuitively reflect the actual effectiveness of the fix methods.

\textbf{Evaluation Settings.}
In our fix experiments, we remain the same $\gamma=0.1$ for random sampling and choose the same key layers. For the threshold, we set $m=1$ to determine $Neun^\uparrow$ and $Neun^\downarrow$. 

\subsection{RQ5 (Effective Fix): How effective is our approach in fixing glitch tokens across different LLMs?}

We compared the performance of the \tool fix algorithm with the rule-based fix method in terms of repaired tokens and repair rate under the same conditions to evaluate the effectiveness of \tool. Table \ref{fix Results} presents the performance comparison of the two methods across different test models.

\begin{table}[t]
    \centering
    \caption{Performance comparison of \tool and Rule-based method on different models.}\vspace{-0.4cm}
    \setlength{\tabcolsep}{2pt}
    \resizebox{\linewidth}{!}{
    \begin{tabular}{l|c|c|c}
         \hline
         \multirow{2}{*}{\textbf{Model}} 
         & \multirow{2}{*}{\textbf{Metric}}
         & \multicolumn{2}{c}{\textbf{Method}} \\ \cline{3-4}
         & & \textbf{Rule-based Fix}  & \textbf{\tool}  \\ \hline
         \multirow{2}{*}{Llama-2-7b-chat} & Repaired Tokens & 3,805 & 4,021 \\
         & Repair Rate & 59.22\% & 62.58\%  \\ \hline
         \multirow{2}{*}{Mistral-7B-Instruct-v0.1} & Repaired Tokens & 359 & 1,045 \\
         & Repair Rate & 12.92\% & 37.60\%  \\ \hline
         \multirow{2}{*}{Qwen-7B-Chat} & Repaired Tokens & 10,645 & 14,765 \\
         & Repair Rate & 34.68\% & 48.11\%  \\ \hline
         \multirow{2}{*}{Gemma-2b-it} & Repaired Tokens & 9,865 & 13,638 \\
         & Repair Rate & 35.28\% & 48.77\%  \\ \hline
         \multirow{2}{*}{Yi-6B-Chat} & Repaired Tokens & 3,390 & 4,317 \\
         & Repair Rate & 41.83\% & 53.26\%  \\ \hline
         \multirow{2}{*}{Average} & Repaired Tokens & 5,613 & 7,758 \\
         & Repair Rate & 36.79\% & 50.06\%  \\ \hline

    \end{tabular}
    }    \vspace{-0.4cm}
    \label{fix Results}
\end{table}

The results indicate that although the rule-based fix method uses fixed $\alpha$ and $\beta$ values ($\alpha=4$ and $\beta=1.5$)  and lacks flexibility, it can still direct the activation or inhibition of neurons associated with glitch tokens, achieving a certain degree of fix. This demonstrates that adjusting neuron activation values to correct the abnormal behavior of glitch tokens is a viable and effective fix strategy. However, due to its lack of specificity, the fix effect of this method is relatively limited. In contrast, \tool precisely calculates $\alpha$ and $\beta$, and selectively adjusts the activation patterns of key feature neurons, achieving an average repair rate of 50.06\% across the five models, outperforming the rule-based method.

\begin{tcolorbox}[boxsep=3pt,left=4pt,right=4pt,top=2pt,bottom=2pt,colback=gray!20,colframe=gray!80!black,title=Answer to RQ5]
\tool has effectively fixed glitch tokens across five LLMs. Compared to the rule-based method, its key improvement is the precise calculation of $\alpha$ and $\beta$, allowing better application to different models.
\end{tcolorbox}

\section{discussion}

\subsection{Hyperparameters Choice of \tool}

\begin{table}
\centering
\caption{Post process time of \tool using various SVM parameter configurations (in seconds) }\vspace{-0.4cm}
\setlength{\tabcolsep}{2pt}
\label{tab:post_process_time}
\resizebox{\linewidth}{!}{%
\begin{tabular}{lcccc}
\toprule
\textbf{Feature Type} & \textbf{\makecell{C=0.1\\degree=2}} & \textbf{\makecell{C=0.5\\degree=2}} & \textbf{\makecell{C=0.5\\degree=3}} & \textbf{\makecell{C=1\\degree=3}} \\
\midrule
Attn\_pattern & 1,413.88 & 1,453.21 & 1,360.12 & 1,357.70 \\
MLP\_gate & 1,407.36 & 1,458.88 & 1,436.23 & 1,445.11 \\
MLP\_data & 1,473.43 & 1,541.10 & 1,579.08 & 1,581.00 \\
Attn\_pattern + MLP\_gate & 1,410.79 & 1,438.57 & 1,444.74 & 1,472.50 \\
Attn\_pattern + MLP\_data & 1,427.43 & 1,479.34 & 1,514.72 & 1,513.41 \\
MLP\_gate + MLP\_data & 1,453.72 & 1,504.36 & 1,546.09 & 1,530.88 \\
Attn\_pattern + MLP\_gate + MLP\_data & 1,444.23 & 1,471.85 & 1,502.46 & 1,500.23 \\
\bottomrule
\end{tabular}
}\vspace{-0.2cm}
\end{table}

\label{subsec:choice}

In this section, an experiment is conducted to illustrate the selection process for the principal features in LLMs and the hyperparameters in SVM. Specifically, the parameters $C$ and $degree$ in the SVM's polynomial kernel require elucidation. The parameter $C$, commonly referred to as the regularization parameter, controls the trade-off between achieving a low error on the training data and minimizing the model complexity for better generalization to new data. A higher value of $C$ tries to fit the training set as well as possible (higher model complexity), while a lower value leads to a model that might not perform as well on the training set but is better at generalizing. On the other hand, $degree$ pertains to the degree of the polynomial kernel function and is crucial for defining the complexity of the decision surface. A higher $degree$ results in more complex decision boundaries, capable of capturing more intricate patterns in the data. However, this also increases the risk of overfitting, particularly in scenarios with noise and limited data samples~\cite{vapnik1995nature,scholkopf2002learning}.

\begin{figure}[htbp]
    \centering
    \resizebox{\linewidth}{!}{
        \includegraphics[width=0.5\columnwidth]{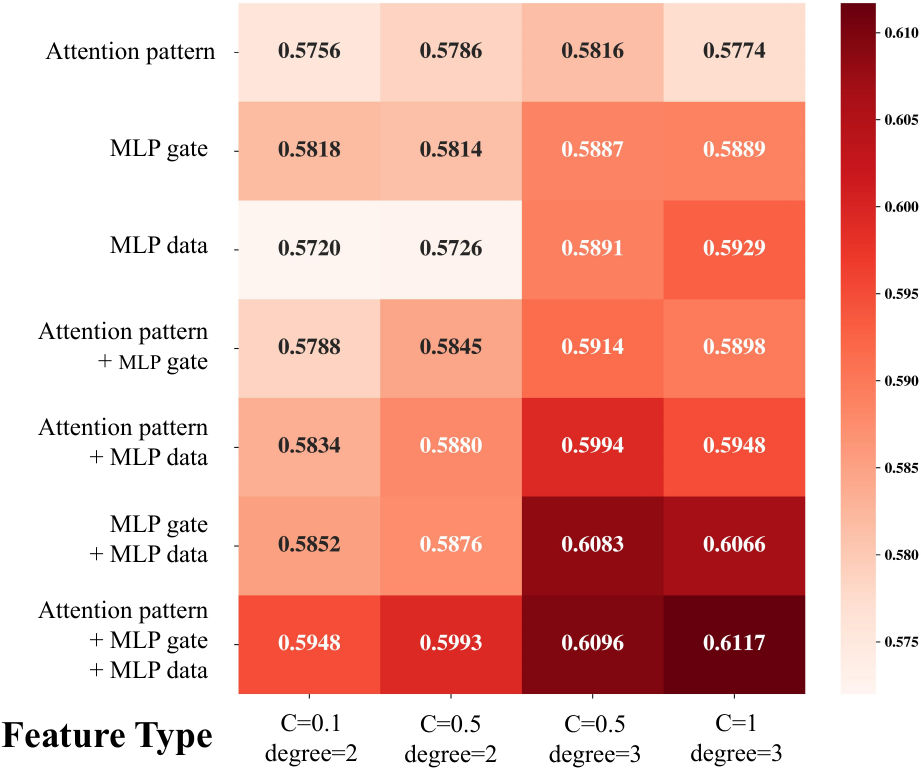}
    }
    \caption{Different Feature Combination Comparison for \tool}
    \label{fig: Feature type comprison}
\end{figure}

As depicted in Table~\ref{tab:post_process_time}, no significant differences are observed in the time consumption across various groups of hyperparameters and features. Therefore, the average F1-score, as illustrated in Figure~\ref{fig: Feature type comprison}, is considered for comparison. The comparison of the F1-score without post-processing is chosen as it more accurately reflects the inherent effectiveness of the different features and hyperparameters. Figure~\ref{fig: Feature type comprison} clearly shows that the hyperparameter group in the lower right corner achieves the highest F1-score of 0.6117 without post-processing, which is noteworthy. Consequently, \( C=1 \) and \( \text{degree}=3 \) are selected as the hyperparameters for SVM. All three features are chosen for the detection process, and two MLP-based features are chosen for the fix process.
\vspace{0.2cm}

We further examine the rationale behind the selection of the hyperparameter $\gamma$, which determines the sampling rate from the model's token vocabulary $V$ for the sample set $S$ in GlitchProber. The parameter $\gamma$ is pivotal in balancing detection accuracy against computational efficiency. Empirical analysis, as shown in Table~\ref{tab:performance_gamma}, demonstrates that a $\gamma$ increment from 0.1 to 0.3 improves recall from 0.6922 to 0.7457 and raises the F1 score from 0.8181 to 0.8543, with a reasonable increase in computational time. However, further increasing $\gamma$ to 0.7, while boosting recall and F1 scores to 0.7812 and 0.8772 respectively, significantly extends processing times to over 100 minutes. Therefore, a $\gamma$ range of 0.1 to 0.3 is recommended, as it optimally balances performance gains with computational efficiency, ensuring that GlitchProber remains practical for operational use.

\begin{table}
\centering
\caption{Model performance across different gamma values on Llama2 model}\vspace{-0.2cm}
\label{tab:performance_gamma}
\resizebox{\linewidth}{!}{
    \begin{tabular}{l|c c c c c}
    \hline
    \textbf{Sampling Rate} & \textbf{$\gamma=0.1$} & \textbf{$\gamma=0.2$} & \textbf{$\gamma=0.3$} & \textbf{$\gamma=0.5$} & \textbf{$\gamma=0.7$} \\
    \hline
    \textbf{Precision} & 100\% & 100\% & 100\% & 100\% & 100\% \\
    \hline
    \textbf{Recall} & 69.22\% & 72.81\% & 74.57\% & 76.65\% & 78.12\% \\
    \hline
    \textbf{F1 Score} & 81.81\% & 84.26\% & 85.43\% & 86.78\% & 87.72\% \\
    \hline
    \textbf{Time} & 61min 38s & 68min 14s & 74min 30s & 86min 52s & 100min 41s \\
    \hline
    \end{tabular}
} 
\end{table}
\vspace{-0.1cm}

\subsection{Rationale of \tool versus Exhaustive Search}  

In the context of exhaustive search mechanisms, LLMs are required to generate, on average, \textbf{twenty} tokens at a zero temperature setting for every individual token processed. This computational method is both intensive and inefficient. Conversely, the implementation of \tool necessitates merely a \textbf{single} forward processing step to capture the intermediate features of the LLM for each token. 
This approach substantially reduces 95\% of redundant operations of those required by the traditional exhaustive search method. 


\subsection{Glitch Token Mitigation By Fine-tuning}

\begin{table}
\centering
\caption{Performance comparison of original and modified model using Finetuning and GlitchProber for Mitigation}\vspace{-0.4cm}
\label{tab:mitigation_impact}
\resizebox{\linewidth}{!}{
\begin{tabular}{lccc}
\hline
\textbf{Dataset} & \textbf{Original Model} & \textbf{GlitchProber} & \textbf{Finetuning} \\
\hline
GSM8K & 0.315 & 0.301 & 0.238 \\
HumanEval pass@1 & 0.129 & 0.103 & 0.009 \\
HumanEval pass@5 & 0.190 & 0.161 & 0.024 \\
MMLU & 0.453 & 0.417 & 0.229 \\
\hline
\end{tabular}
}\vspace{-0.4cm}
\end{table}
\vspace{-0.1cm}

\tool adaptively modifies the process of model calculation without altering the model parameters, thereby minimizing the mitigation impact on model performance and preserving the basic abilities of LLMs. In contrast, we also construct a dataset with Q\&A for the repetition task and attempt to mitigate the glitch token phenomenon by fine-tuning LLMs. However, compared with \tool, fine-tuning LLMs alters the parameters of the model, potentially compromising its basic abilities. For example, we fine-tune the Llama-2-7b-chat with a dataset containing 3,000 Q\&A pairs of repetition tasks. To evaluate the model's basic skills, we use three widely accepted datasets namely \texttt{GSM8K}~\cite{cobbe2021gsm8k}, \texttt{HumanEval}~\cite{chen2021evaluating}, and \texttt{MMLU}~\cite{hendrycks2020measuring}. 

Detailed results presented in Table~\ref{tab:mitigation_impact} indicate that the model's ability in code writing and solving math problems post \tool is comparable with the original model. and significantly better compared to the fine-tuned model, which notably diminished the basic abilities of original model.

\subsection{Threats to Validity}

\textbf{Internal.} Our primary concern involves the selection and computation of hyperparameters in both the detection and fixing phases of \tool. For the detection phase, we detail the rationale behind our choices through an experiment described in Section~\ref{subsec:choice}. Various feature types and hyperparameters for the SVM consistently outperform the established baselines. In the fixing phase, the linear computation of $\alpha$ and $\beta$ proves untenable. Enhanced discussion and manipulation of the activation value are recommended as future research directions.

\noindent\textbf{External.} Threats are associated with our experimental framework. For the performance of glitch tokens in intermediate layers, we experiment on three LLMs with different parameters and vocabulary sizes. Based on these findings, \tool experiment on five different LLMs, showing its generalizability. Furthermore, as \tool is required to access the intermediate data of the LLM, \tool is only applicable to open-source LLMs. The transferability of \tool from open-source LLMs to closed-source LLMs like GPT-4 could further be explored.

\vspace{-0.2cm}
\section{conclusion}


In this work, we reveal that glitch tokens trigger abnormal activation characteristics in the model's attention patterns and MLP status through systematic empirical study. Inspired by this, we propose methods for detecting and fixing glitch tokens. \tool employs a sampling strategy, extracting features from attention patterns and MLP status, and achieves rapid screening of the vocabulary through PCA dimensionality reduction and SVM classification. Experiments on LLMs demonstrate that \tool saves 40\% of the time compared to existing methods while achieving higher accuracy. Another important contribution is our strategy to fix glitch tokens by adjusting the activation values of the model's intermediate layers. Experiments on LLMs confirm the effectiveness of this fix strategy, and the average repair rate of \tool was improved by 13.27\% compared with the baseline method.

\section*{Acknowledgement}
This work was supported by the National NSF of China (grants No.62302176, No.62072046, 62302181), the Key R\&D Program of Hubei Province~(2023BAB017, 2023BAB079), and the Knowledge Innovation Program of Wuhan-Basic Research (2022010801010083).

\bibliographystyle{ACM-Reference-Format}
\bibliography{9-paper}

\end{document}